# DESIGN OF HAZARD BASED MODEL AND COLLISION AVOIDANCE SYSTEM

Md Faysal Kabir[1], Sahadev Roy[2]

[12]Department of ECE, NIT Arunachal Pradesh, Yupia, India

Email: [1]juel.kabir@gmail.com, [2]sahadevroy@nitap.ac.in

**ABSTRACT:** The primary goal of this paper is to examine the motorcyclists' activities during the overtaking period, as well as to develop a model of total overtaking time. For the experimental study, instrumented motorcycles were used to collect data and design an overall overtaking period model. The possibility of death during an attempt to overtake is maximum for people walking on rural roads, mainly identified by a scarcity of pathways and higher speeds of vehicles. It is important to recognize and prototype driver actions during the overtaking moves to set up collision avoidance strategies in order to prevent these collisions, such that device adjustments are acceptable as they occur beyond the comfort zone of the drivers. The purpose of this research is to address both vehicle overtaking movements along urban roads and to develop a collision-avoidance system. This research, based on tests performed and instrumented driving information, may lead to the discovery of advanced driver assistance systems, analyzing driver behavior during overtaking. A total of 500 overtaking movements were registered with 50 motorcycles set up with a high-resolution camera and GPS system implemented by 50 professional bikers in India in an undivided one-way road. A technique was developed to collect data explaining the actions of the motorcyclists, based on video and GPS analyses. The overall overtaking period was designed using a risk-based model, which indicates the span of the overtaking based on several coefficients. The proposed model is useful to analyze the behavior during overtaking moves, as well as to develop road and vehicle safety systems to reduce the chances of accidents.

## I. INTRODUCTION

The motorcyclists attempts the overtaking activity is a major reason of automobile accidents. Although, only a few number of research work undergoes regarding with such subject while there are all most no research about the calculation about the overall overtaking period, that is the duration during which serious threats are obvious. Two-wheelers are the country's most sensitive and unsecured vehicles on the road. The overall number of deaths worldwide has already reached the 1.35 million mark in 2018 [1], with India responsible for all most 11 per cent of the world 's total accident-related deaths. In 2018, as per the report of World Road Statistics India ranks number one among 199 countries with 1.5 lakhs of deaths [2] because of road accidents. Among these 1.5 lakh of deaths, 40% are due to the two or three wheelers and the Pedestrians are responsible for 10 per cent of all traffic deaths [3].

The proportion of 2-wheeler users crash fatalities rose from 31.5% (46,070 people) to 34.8% (52,500 people), from the year 2015-16 (Road Accidents in India – 2016 and 2015 by Government of India, Ministry of Road Transport & Highways). Every year, in road accidents the working age population loses their lives. Preventable loss of these younger generations produces significant productivity lack for the country. In 2016, the deaths from road accidents indicate that the 18-35-year efficient age group reported for a high 46.3 percent share (69,851 people) and the 18-45-year age group accounted for 68.6 percent (1,03,409 people) of the total deaths from road accidents (Road Accidents in India – 2016) . Gender based analysis of deaths of males and females in road traffic accidents shows that the overall proportion of men and women killed in the 2016 calendar year was 1,27,453 (84.5%) and 23,332 (15.5 %) in total road accidents, respectively (Road Accidents in India – 2016).

The two wheeler's are the main components of Indian traffic, and they are responsible for the maximum number of deaths and injuries (as shown in Table 1), as they come in direct contact during a crash with the impacted object [4]. The main concern with the health of motorcyclists is attributable in theory to the variation in the driving techniques  between the other vehicles and the two wheelers [5].

Several researchers have studied the most dangerous activities impacting the health of motorcyclists. They found maturity, sexual category and expertise are the parameters that affects [6] the behavior of the motorcyclist. Maturity performs quite a significant role than expertise, as injuries are correlated with tendency of violating the rules and breaking secure driving guidelines. Younger male bikers are the most vulnerable to be a victim in crashes, and the most of the road crashes was found among motorcyclists in the 18-35 year age group [7].

Information about the age group, gender, nature of road, incident date, type of vehicle involved, severity of injuries and type of road users must be collected in order to design a preventive system for motorcyclists' overtaking maneuvers. Overtaking is one of the most risky and complex turns on the undivided roads as the vehicle attempting to overtake the in front vehicle has to pass through the opposite lane with traffic ahead approaching from opposite direction [8].

Several factors underpin this analysis. There are few research studies that concentrate on riders' actions while driving or overtaking, but the preventive steps to be taken in order to minimize the number of incidents are limited and scarce in the literature [9], [10], [11]. Recent research on the motorcyclists' hazard-based overtaking period model, calculates the time required in overtaking [12], [13] but no evidence has been found of how to recover from this accident . The main objective of this study is therefore to quantify the vehicle overtaking period and the preventive measures that can reduce the number of accidents caused by overtaking.

When the driver attempts to overtake the vehicle in the front of his/her bike, the bike rider tries to adjust his/her angle of motion. During this overtaking process, the bike rider crosses the centre - line and creates a slight gap from the vehicle that the rider wants to overtake. In this overtaking cycle the bike rider is absolutely unaware about any information about the incoming traffic from the opposite lane, which can in turn cause a deadly collision.

**Table 1: Comparison between number of accidents in 2015, 2016, 2017 and 2018**

| Year | Total no of accidents | % change in total number of accidents | 18-35 | 18-45 | Over-speeding | Rural area | Urban Area | Two wheelers |
|---|---|---|---|---|---|---|---|---|
| 2015 | 5,01,423 | 2.46 | 42.5 | 64.8 | 77.1 | 53.8 | 46.2 | 31.5 |
| 2016 | 4,80,652 | -4.14 | 46.3 | 68.6 | 66.5 | 54.9 | 45.1 | 34.8 |
| 2017 | 4,64,910 | -3.28 | 49.9 | 72.1 | 70.4 | 57.9 | 42.1 | 33.9 |
| 2018 | 4,67,044 | 0.46 | 48.0 | 69.6 | 66.5 | 59.1 | 40.8 | 35.2 |

In order to avoid these collisions the motorcyclists must have some information about the incoming traffic from the opposite side and that inspire this work to develop a safety system for the bike riders that can able to save the valuable life of young generation of our country. To design the safety system the first step is to calculate the duration of overtaking of the in front vehicle then by analyzing the speed of the motorcycle and by the time taken in overtaking the exact system can be developed.

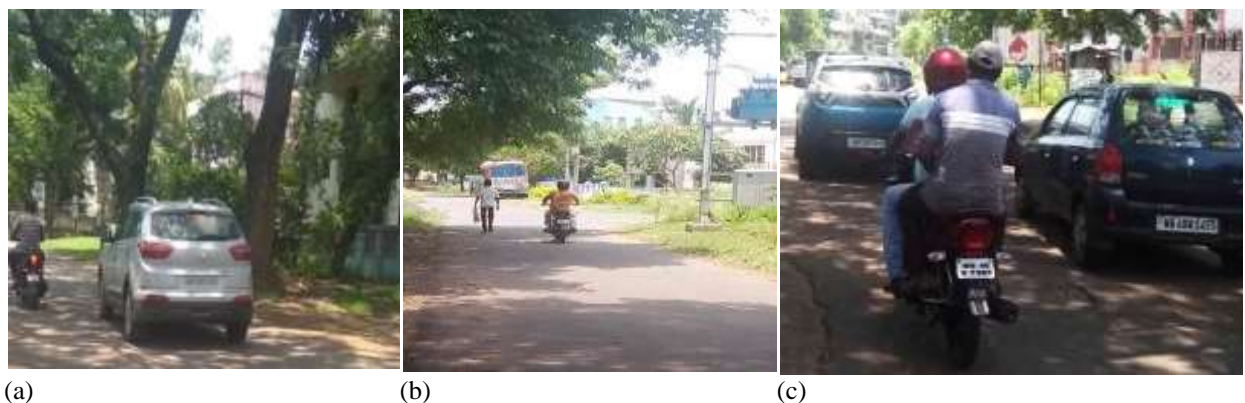

(a) (b) (c)

**Fig .1. Images taken during the experiment: (a) vehicle overtaken by the motorcyclist, (b) pedestrian overtaking by the motorcyclist, (c) mixed traffic motorcyclist**

## II. METHODOLOGY

The starting and finishing positions of the overtaking operation were identified through the mapping of the longitudinal location pattern followed by the motorcycle driver. On the basis of the position of the bike rider, from the beginning to the end of the overtaking cycle, the entire procedure is divided into 5 phases, as shown in Figure 2,
- period 1: It is the period when the rider starts the overtaking procedure and moves towards the centre line by increasing the speed of the vehicle,
- period 2: The motorcyclists cross the centre line and occupies the opposite lane
- period 3: In this period the bike crosses the Car 1 by creating some distance through the opposite lane and again coming back towards the centre line
- period 4: Now, in phase 4, the motorcyclists comes back to his correct lane of riding and continues to move centerline to go back to the right lane
- period 5: This is the phase where the bike rider again reduces the speed of the vehicle and continue to move with the velocity with the same speed before overtaking.

In period 3 and in period 5 there are chances of collision with Car 2 and Bike 2 respectively, as during the overtaking procedure the bike rider increases the speed of his/her vehicle.

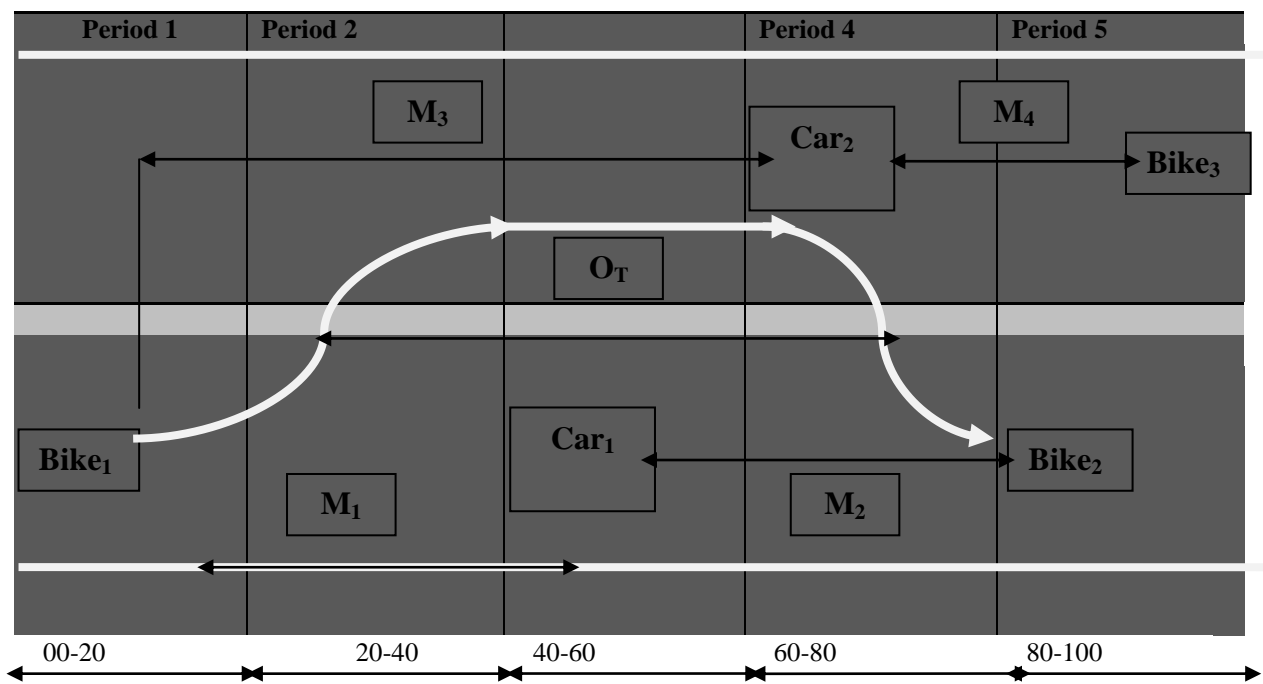

ii.     Road geometry for the calculation of overtaking duration
In this work, the parameters that has been used for the study of overtaking movements of the motorcyclists are as follows:

$A_1$- Velocity of $Bike_1$ before starting the overtaking procedure,
$A_{12}$- Velocity of the $Bike_1$ when it crosses the centre line in order the overtake $Car_1$(gradually increasing in nature), $A_{11}$- Velocity of the $Bike_1$ when it again comes back to its right lane, B- Velocity of $Car_1$, C- Velocity of $Bike_2$, D- Velocity of $Car_2$, E- Velocity of $Bike_3$, $M_1$= Distance between $Bike_1$ and $Car_1$, $M_2$= Distance between $Car_1$ and $Bike_2$, $M_3$=Distance between $Bike_1$ and $Car_2$, $M_4$= Distance between $Car_1$ and $Bike_2$.

Some previous researchers examined that the calculation of mutual distance (distance between Car1 and Bike1 during overtaking) and the distance with the in front vehicle is only possible if both the in front vehicle and the oncoming traffic from the opposite lane are visible to the motorcyclists [14], [15], [16]. Design of hazard based model by assuming the road is straight and all the vehicles are visible to the motorcyclist increases the chances of car crash during overtaking. In this proposed work , the distance between each vehicle present in the road (obviously which vehicle are in motion) is being calculated with the help of Satellite. So, this process does not require any kind of direct visibility of the incoming traffic from the opposite lane, for the bike rider (Bike1).
The width of the road is 10 m which composed of two 4 m wide lanes. The orientation level was negligible and the maximum specified velocity was 50 km/hr. The selection of this road is based on the Indian road safety plan review, as the report identifies this road among the most dangerous one, for the ctwo wheelers (Figure 1). Young male bikers within ages of 21 and 30, With a median age of 25 years, took part in this studies. Young male bike riders were chosen for this analysis as this category of riders is the most vulnerable to incident [17]. The eligibility criteria for the motorcyclists to participate in this experiment was at least 5 years of riding experience and the rate of bike use should also be at least 4 to 6 days a week. The motorcycles took part in the experiment are equipped with a high-resolution camera (Go Pro Hero3) and a GPS feature. A high-resolution camera is mounted in front of the bikes, which can able to take 30 images per second, and the GPS data provides positioning information about the in front vehicle as well as oncoming traffic from the opposite lane.

### III. DATA ANALYSIS

Data processing was achieved by evaluating the overtaking procedure variables, collected from both Video and GPS data (Table 2). An initial process of Video and GPS data synchronization has done through the data collected from each participant's driving test using the Dashware Applications. This software allows visualization of the captured video and synchronization of data generated by GPS system. Both the video and GPS data were gathered and coordinated to assess and categorize overtaking experiences (on the basis of total vehicle crossed during the overtaking process), the occupation of the oncoming lane and the termination factors. Once the occurrence sets of images were gathered for all the overtaking exercise, the technique mentioned below was being used to evaluate the key parameters. This is designed to evaluate width of the lane by Psarianos et al., 2001 [18], by synchronization of visual information collected from the video and GPS, and later upgraded to achieve inter-vehicular distances as well as velocities. The findings indicate that certain linear and mutual distances can be measured from the relationships below(using Figure 3), as shown in Equation (1):

$$Z = (y_F - y_G)/cY_I \quad (1)$$

Where, c represents camera constant, Z represents linear distance, $Y_F$ indicate a constant which implies the focus length of the camera within the co-ordinate F and F', $Y_G$ indicates blind point of image y and $Y_I$ is the initial height at which camera is placed above ground level, $\Delta X_F$ represent the mutual distance between the points F and F'.

**Table. 1. Explanatory descriptions of the overtaking movement parameters**

| Variables of the overtaking maneuver | Explanatory information | Values of the variables | | |
|---|---|---|---|---|
| | | N | Mean | St.dev |
| N | Total number of vehicle passed during the overtaking period | 500 | 6 | 0.4 |
| $T_{total}$ | Total period of overtaking | 100 | 6.5 | 2.1 |
| $TP_1$ | First time period | 20 | 7.5 | 1.0 |
| $TP_2$ | Second time period | 20 | 8.2 | 1.8 |
| $TP_3$ | Third time period | 20 | 6.5 | 0.9 |
| $TP_4$ | Fourth time period | 20 | 5.9 | 0.5 |
| Dtotal | Total overtake distance | 100 | 105.5 | 40.4 |
| $DP_1$ | Distance in 1st period | 20 | 8.5 | 17.7 |
| $DP_2$ | Distance in 2nd period | 20 | 8.1 | 20.5 |
| $DP_3$ | Distance in 3rd period | 20 | 7.6 | 24.1 |
| $DP_4$ | Distance in 4th period | 20 | 7.0 | 18.7 |
| $M_1$ | Primary distance | 30 | 8.3 | 5.0 |
| $M_2$ | Ultimate distance | 30 | 6.9 | 7.2 |
| M | Mutual distance | 6 | 5.5 | 0.5 |
| $A_1$ | Velocity of Bike1 before starting the overtaking procedure | 500 | 281 | 10.5 |
| $A_{12}$ | Velocity of the $Bike_1$ when it crosses the centre line in order the overtake $Car_1$ | 500 | 300 | 9.9 |
| $A_{11}$ | Velocity of the $Bike_1$ when it again comes back to its right lane | 500 | 234.5 | 12.3 |
| dAB | Speed difference between $Bike_1$ and $Car_1$ before overtaking | 500 | 46.5 | 8.4 |
| $M_3$ | Mutual Distance between $Bike_1$ and $Car_2$ | 70 | 8.4 | 5.6 |
| $M_4$ | Mutual Distance between $Car_1$ and $Bike_2$ | 30 | 6.5 | 4.9 |

Using Equation (1), certain linear distances can be measured, such as $M_1$, $M_2$, $M_3$, and $M_4$. By combining video with GPS data, the speed of the motorcycle could be obtained at every instant of the overtaking process. The velocity of past vehicle (B) and the approaching car (D) could be determined by the speed of the motorcycle ($A_1$, gathered from GPS data) and the gap in between vehicles (Z); with the following equation:

$$B = (A_1(t-1) + A_1(t))/2 + (Z(t) - Z(t-1))/\Delta t \qquad (2)$$

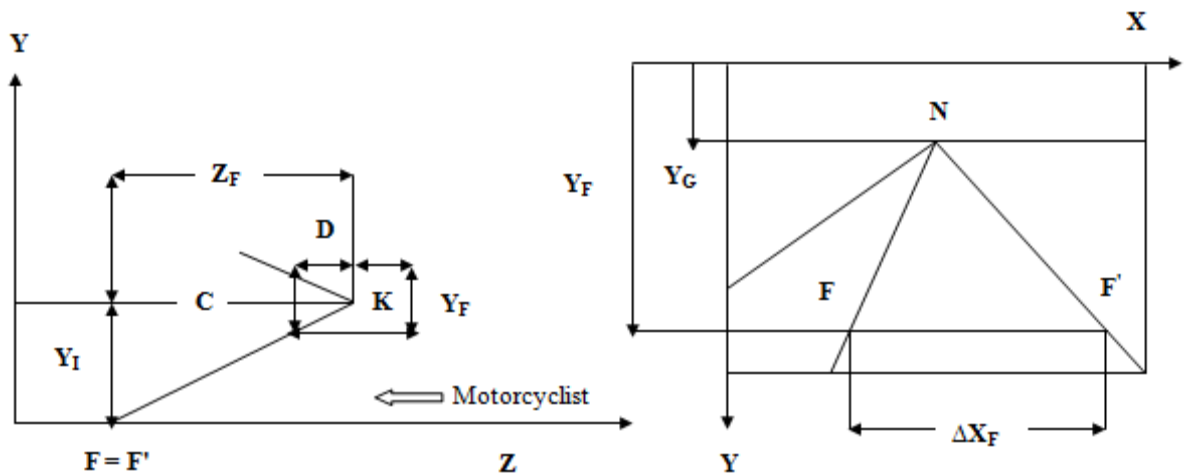

iii.　　Image acquisition geometry

In order to test the consistency of Equation (2), item of known dimension ($Z_{mj}$) used during the calibration process positioned at 6 separate measured distances, from 5 m to 30 m. Then, the equation (1) was used to measure those 6 spans. The mean absolute percentage error (MAPE) for all the calibration processes was given by equation (3):

$$\text{MAPE} = \frac{1}{n}\sum_1^n \frac{100}{p} \sum_{j=1}^n \left|\frac{Z_{cj} - Z_{aj}}{Z_{mj}}\right| = 4.23\% \qquad (3)$$

where,

$Z_{cj}$ = calculated distance, $Z_{ej}$ = anticipated distance, p = no. repetitive investigations (i.e., 6), and n = total number of calibration, (i.e., 30) (total number of participator).

Similarly, with the help of object width, the MAPE obtained for the mutual distances was 2.26 percent.

It is observed from this study that the method was accurate where the distance amid the camera and the part of importance are in the range of 5 m to 30 m. In the case of longer distances the degree of precision must be checked.

Explanatory details of different overtaking procedure variables are given in Table 2. Only 320 cases were reported among the 500 cases of overtaking when the biker enters the opposite lane in order to overtake. The mean overall overtaking time and overtaking distance is 6.5s and 115 m, respectively. Considering just the invasion maneuvers (320), the overtaking was mainly conducted in the three-period, with a mean time span spent on the opposite lane, 4 s (with a small amount of time during period 2 + Entire period 3 + little time in period 4).

Unfortunately, no research has been found that measures the motorcyclists' occupation time in the opposite lane during overtaking [19], [20]. It is important to note that previous research found higher values for this variable when four-wheeled vehicles were overtaken. Just 30 overtaking turns can be estimated for ultimate distance (M2) and mutual distance between Car1 and Bike2 ($M_4$).

This is significant to mention that the average value of the mutual distance M, and the final distance $M_2$, were 1.5 m and 9.2 m respectively in comparison with the result of 1.31m, achieved by Vlahogianni [21], who investigated the behavior of motorcyclists in urban areas, similarly higher in case of ultimate distance higher values were achieved for four-wheeled vehicles (19.9 m), as stated by Polus [22].

# 1. DESIGN OF HAZARD BASED MODEL AND COLLISION AVOIDANCE SYSTEM

The overall overtaking duration was modeled using a hazard-based model, which seems to be the most appropriate tool to study the overtaking duration variables. This research suggests a hazard-based duration model to characterize the behavior of bikers during overtaking while at the same time this model is useful for calculating the time span for any particular state of riding [23], which in turn becomes fruitful in the design of collisions avoiding system. Additionally, it is capable of using sensor-based knowledge to enhance the measurement accuracy [24].

The hazard based model is useful in the field of transport to analyze time-related incidents such as probability of road accidents, travel activity etc. Rather than estimating the overtaking period, this model finds its applications in many areas such as, in the industries, in medical science, as well as social sciences too [25]. Recently, few researcher have conducted hazard-based study to examine pedestrian behavior during the signalized crossing at the intersections [26].

Yang et al . proposed a hazard-based duration model to analyze cyclists and electric bike riders' behavior during crossing at signalized intersections, in China [27]. In their research, David A. Hensher et al. explained the significance and implementation of the hazard-based duration model in transport analysis [28]. Bella et al. developed a hazard-based duration model for the total overtaking period measurement [29]. Gowri Asaithambi et al. examined vehicle overtaking characteristics on undivided roads in conditions of mixed traffic [30], Haque et al. modeled driver response times, interrupted by cell phone conversations [31].

Survival function (S(t)) represents the probability of survival can be expressed mathematically as shown in Equation (4):

$$S(t) = P(T \geq t) = 1 - F(t) \quad (4)$$

where, T represents a non- negative variable which indicates any event (like overtaking), the function $P(T \geq t)$ describes that the probability of survival T should be greater than or equal to t, and F(t) a cumulative probability function.

Probability of density (f(t)) and the survival function (S(t)) is related with each other as shown in Equation 5:

$$S(t) = P(T \geq t) = \int_t^\infty f(t)dt \quad (5)$$

The hazard function h(t) can be obtained by combining both Equation 4&5 and can be expressed mathematically as shown in Equation 6:

$$h(t) = \lim_{\Delta t \to 0} \frac{P(t \leq T < t + \Delta t \, T \geq t)}{\Delta t} \quad (6)$$

Where h(t) indicates the rate of immediate failure based upon the condition of T = t

Hazard function (h(t)) can be also be expressed in terms of S(t) and f(t), as shown in Equation 7:

$$h(t) = \frac{f(t)}{S(t)} = -\frac{d}{dt} \log S(t) \quad (7)$$

The preference behind the AFT model is that this model makes statistics easier to understand, because the calculated parameters analyzes the corresponding effect of a variable on the average life expectancy [32].
Linear form of AFT model is shown in Equation (8):

$$\ln T = \beta X + \varepsilon \quad (8)$$

Where, ε denotes percentage of error

Survival function plays an important role in the statistical approach , in order to select a particular form of distributional function between exponential, log-logistics, Weibull, and gamma distribution. Although among all these distributional function the log-logistics function perfectly matched for the calculation of overtaking duration [33].
Two variables are used to construct the log-logistic function, is shown in Equation (9):

$$\gamma = [\exp(-\beta X)] \quad (9)$$

where, γ = Stretching capacity of the distribution function, exp(βX) = Determinant of AFT model

The biker begins to raise the speed of the bike from the point of crossing the center line and entering the opposite lane, this point of junction is known as the point of intersection or point of inflection. The point of intersection can be expressed mathematically, as shown in Equation (10):

$$\tau^* = \frac{(1/\gamma - 1)^\gamma}{\gamma} \quad (10)$$

The survival function (S(t)), and the hazard function (h(t)) for the log-logistics model can be expressed as shown in Equation (11) and Equation (12) respectively:

$$S(t) = \frac{1}{1 + \exp(-\beta X)t^{1/\gamma}} \quad (11)$$

$$h(t) = \frac{\exp(-\beta X)\frac{1}{\gamma} * t^{1/\gamma - 1}}{\gamma(1 + \exp(-\beta X)t^{1/\gamma})} \quad (12)$$

**IV. RESULT ANALYSIS AND VALIDATION**

The AFT models were developed using the statistical software STATA, version 13.0. Only the variables in a first phase disponible for a total of 500 overtaking were considered as candidates for model development. There were several attempts to adjust the determinants to identify those that were statistically relevant make the model significant, and optimize it. Major AFT model variables required to analyze the biker's overall overtaking time are based on the log-logistics distribution function, as shown in Table 3. The value of the parameter of the scale is obtained as 0.10, which implies that overtaking is possible until the end of the intersection point and then exponentially reduced to the original lane of travel. The probability of survival and the hazard-based function are determined using Equation 11 and Equation 12 respectively and the mean determinant values are plotted in Figure 4.

The interesting fact is that in this method the obtained value of the overtaking duration is of 10s, which is almost equal to that founded by Bella et al. (8s). The slight variance in the result obtained in this analysis with other studies may be due to the observation of more number of overtaking turns or may be due to the difference in road structure between other countries and India, and may also be due to the variation in actions of motorcyclists from different countries.

The model includes statistical modeling of major 5 parameter, which mostly influences the overall period of the overtaking. The sign of the variable dAB (represents variable speed), is negative as the biker accelerates the vehicle's speed after crossing the intersection point which transforms the variable value into negative. Sign of the determinants of log-logistic function is negative as the motorcyclist (Bike1) accelerates the bike's speed during overtaking and drives at a higher speed than the overtaken vehicle (Car1).

**Table. 2.　　Table for variables of the Log-logistics function**

| Name of Variable | β | SE | Linear distance (Z) | P value | Exp(β) | 95% CI | |
|---|---|---|---|---|---|---|---|
| | | | | | | Lower | Upper |
| UD | 0.027 | 0.007 | 8.65 | 0 | 1.34 | 0.009 | 0.036 |
| PD | 0.049 | 0.013 | 10.5 | 0 | 1.35 | 0.056 | 0.089 |
| dAB | -0.053 | 0.012 | -9.25 | 0 | 1.26 | -0.015 | -0.008 |
| Multiple | 0.463 | 0.097 | 7.58 | 0 | 0.82 | | |
| cons | 2.589 | 0.086 | 6.54 | 0 | 2.57 | 0.248 | 0.759 |
| $\gamma$ | 0.253 | 0.069 | - | - | - | 3.547 | 3.773 |
| Log-logistics at convergence | 10.95 | - | - | - | - | - | - |
| Log-logistics at convergence | -150.52 | - | - | - | - | - | - |
| No of observation | 500 | - | - | - | - | - | - |

The fact must be noticed that in order to understand the impact of any particular determinant on survival period, every small difference in the determinant magnitude must be converted into hundred percent using equation exp(-βX). Thus, when the biker tried to perform more than one overtaking turn at a same time, then the requirement of total overtaken time will also raise up to 35 percent, which implies that the biker consumes more time in the case of multiple overtaking and hence increases the chances of collision. Increment of per meter of primary distance ($M_1$) takes 5 per cent more time than the overall overtaking duration, as the biker has to travel more distance to overtake the in front car, which in effect decreases the chances of collision.

Increase in the motorcyclist's speed of 1 km per hour will result in requirement of 4 per cent less time in the total overtaking time. The total overtaking duration increases by 2 percent per meter of increase in the ultimate distance, as motorcyclists gain more space to return to the original lane of travel. Similarly, the increase of 1 m of distance of $M_3$ increases the overall duration of the overtaking by 13%, as the biker gets more time to ride in the opposite lane, resulting in less chances of collision.

We must also consider the specific significance of each parameter of the designed model, to understand the real impact on survival times. From the model, it appears that certain variables, such as the distance between Bike1 and Car2 ($M_3$), had a significant effect on the total overtaking duration, which is equal to 13 percent, with respect to the speed difference (dAB) that is equal to 4 percent. Among these parameters $M_3$ has significant impact on the result, as in this case the overall overtaking duration is increased to 13 percent, which is higher than the impacts of other parameters.

It should be noted that, if we evaluate the average value of $M_3$, it would be approximately equal to 7 m, while the average value of dAB is of 20.3 km/h, hence, the average impact of $M_3$ is higher ($13 \times 7 = 91\%$) in comparison to the average impact of dAB ($20.3 \times 4 = 81.2\%$) on the result.

The analysis therefore concluded that the average impacts of $M_3$ (Figure 4 (c)) and primary distance (Figure 4 (b)) and dAB (Figure 4 (a)) have significant effects on the result (as shown in Figure 4) while the ultimate distance and $M_4$ have minor effects on the result.

Impacts on survival time by major determinants are shown in Table 3. As discussed earlier, in order to understand the real impacts of the determinants, we had to convert each unit of result to 100%, this principle is also applied in the graph plotting. The minimum and maximum obtained result with the average of these two values is being considered to convert the result into 100 per cent. If the speed of the Bike1 ($A_{12}$) is increased over the speed of the Car1 (B), then the overtaking duration will be reduced while all other variable like $M_3$, $M_1$, $M_4$, $O_T$ will significantly boost the overtaking duration. In the case of dAB, if the minimum value of dAB is 5.7 km/hr (20th percentile), then the likelihood that the overtaking process would last longer than 10 s was 55 percent, while, for the maximum value of dAB, i.e. 34.9 km/hr (80th percentile), the likelihood that the overtaking process would last longer than 10 s was reduced to 7 percent. Likewise, for the average dAB value (20.3 km/hr) probability was 27 percent to last the overtaking cycle. Similarly, the effects of all other variables on the survival function were achieved.

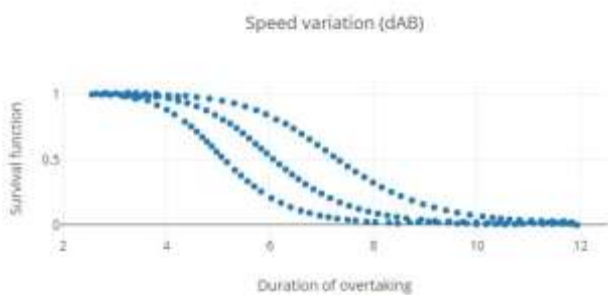

(a)

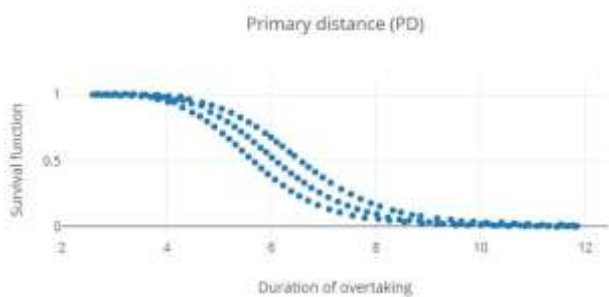

(b)

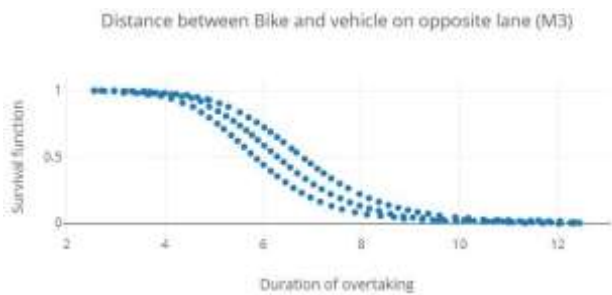

(c)

iv.     Effects of determinants on the survival function: (a) Speed variation (dAB), (b) Primary distance (PD), (c) Distance between bike and the vehicle on the opposite lane

(upper lines are drawn with 20th percentile of the determinants, lower lines with 80th percentile and middle line with average values)

One flow graph that describes the collision avoidance system as shown in Figure 5 has been modeled over here. This system is designed according to the conditions of road traffic shown in Figure 2. The overall overtaking duration obtained from this analysis is 6.5s, and the minimum safe overtaking distance obtained as 115 m, so the device must have a threshold value of 6.5s and 115 m. The inter-vehicle distance (that is moving) can be obtained from the GPS system and a sensor can be developed with the help of these data which can continuously track the speed and position of all traffic present on the road and assure the rider if overtaking is possible or not. This system can save lives, and provides a safe journey for the rides.

After analyzing all the survival and hazard function variables, the overtaking duration obtained as 6.5s, and the minimum distance required to execute an overtaking turn is obtained as 115 m. The log-logistics distribution function fits perfectly into the overtaking duration modeling. The finding achieved implies that the likelihood of making an overtaking turn after the point of intersection is higher that continues until the end of the $O_T$ area after this exponentially decreases to zero.

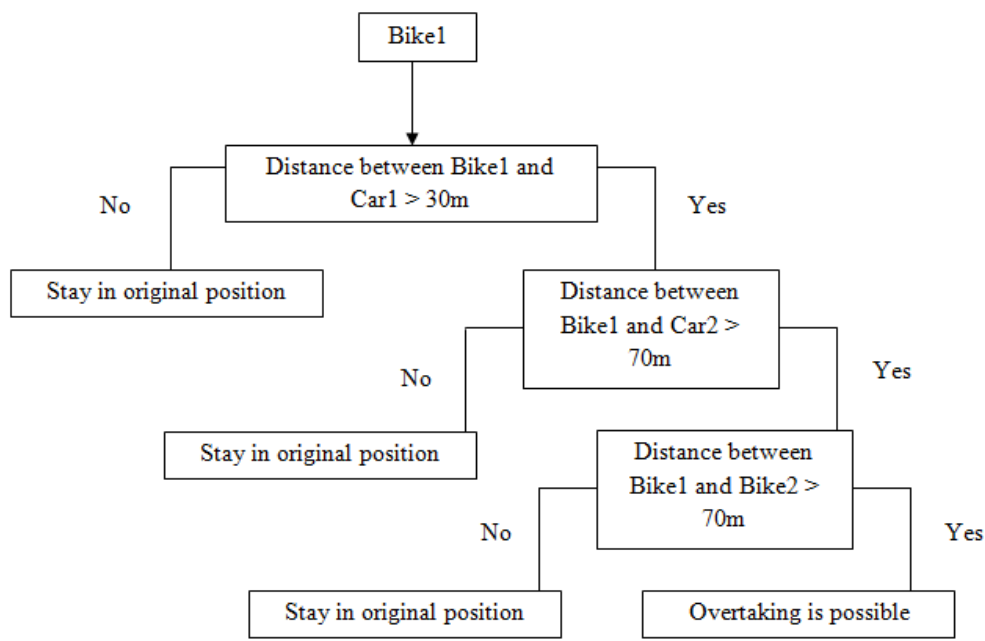

v.     Flow graph for the design of collision avoidance system

From the result obtained, it is obvious that determinants like primary distance (PD), ultimate distance (UD), distance from Bike1 and Car2 ($M_3$), speed difference between Bike1 and Car1 (dAB) etc. has significant impacts on the model. However, the primary distance, gap between the motorcycle and the oncoming traffic from the opposite lane and the speed difference between the biker and the overtaken vehicle has greater impact on the result compared to the effects of the ultimate distance and the $M_4$.

## V. CONCLUSION

This paper focuses on modeling motorcyclists' overall overtaking period on a two lane urban road to avoid collision. A field experiment with instrumented motorcycles has been carried out to design the model. In order to characterize the behavior of motorcyclists during overtaking, the methodology analyzes the data obtained from GPS and high resolution camera, with the help of survival analysis function. Development of the collision avoidance system is possible with the help of obtained results total time taken in overtaking, effects of speed and distance in overtaking process etc. For riders, the collision avoidance system would act as a supporting platform that can assure the motorcyclist that if an overtaking turn is possible, or not. This advance protection system compares the time required between the vehicle on the opposite lane as well as the in front vehicle, also calculates the overtaking duration based on the current conditions of riding. The experimental results shows that the proposed experimental results are successfully validate our proposed model.